\documentclass[nonacm=True, format=sigconf]{acmart}
\usepackage[utf8]{inputenc}

\usepackage{amsmath}
\usepackage{graphicx}
\usepackage[subrefformat=parens,skip=0pt, font=small]{subcaption}
\usepackage{siunitx}
\usepackage{enumitem}
\usepackage{bm}
\usepackage[ruled,vlined]{algorithm2e}
\usepackage{cleveref}

\newcommand{\set}[1]{\mathcal{#1}}
\newcommand{\mat}[1]{\bm{#1}}

\newcommand{\given}{\colon}
\newcommand{\R}{\mathbb{R}}
\newcommand{\norm}[1]{\left\|#1\right\|}

\DeclareMathOperator*{\argmin}{arg\,min}

\newcommand{\ltg}{{\textit{local2global}}}

\title{Local2Global: Scaling global representation learning on graphs via local training}
\acmConference{}{}{}

\author{Lucas G. S. Jeub}
\affiliation{%
\institution{The Alan Turing Institute}
\country{}
}
\email{lucasjeub@gmail.com}

\author{Giovanni Colavizza}
\affiliation{%
\institution{University of Amsterdam}
\country{}
}
\email{g.colavizza@uva.nl}

\author{Xiaowen Dong}
\affiliation{%
\institution{University of Oxford}
\country{}
}
\email{xdong@robots.ox.ac.uk}

\author{Marya Bazzi}
\affiliation{%
\institution{University of Warwick} \institution{The Alan Turing Institute}
\country{}
}
\email{marya.bazzi@warwick.ac.uk}

\author{Mihai Cucuringu}
\affiliation{%
\institution{University of Oxford} \institution{The Alan Turing Institute}
\country{}
}
\email{mihai.cucuringu@stats.ox.ac.uk}

\date{\today}
\keywords{scalable graph embedding, distributed training, group synchronization}

\begin{document}

\begin{abstract}
We propose a decentralised ``{\ltg}'' approach to graph representation learning, that one can a-priori use to scale any embedding technique. Our {\ltg} approach proceeds by first dividing the input graph into overlapping subgraphs (or ``\textit{patches}'') and training local representations for each patch independently. 
In a second step, we combine the local representations into a globally consistent representation by estimating the set of rigid motions that best align the local representations using information from the patch overlaps, via group synchronization.  A key distinguishing feature of {\ltg} relative to existing work is that patches are trained independently without the need for the often costly parameter synchronisation during distributed training. This allows {\ltg} to scale to large-scale industrial applications, where the input graph may not even fit into memory and may be stored in a distributed manner. 
Preliminary results on medium-scale data sets (up to $\sim$7K nodes and $\sim$200K edges) are promising, with a graph reconstruction performance for {\ltg} that is comparable to that of globally trained embeddings. A thorough evaluation of {\ltg} on large scale data and applications to downstream tasks, such as node classification and link prediction, constitutes ongoing work. 
\end{abstract}

\maketitle

\section{Introduction}
\label{introduction}

The application of deep learning on graphs, or Graph Neural Networks (GNNs), has recently gained considerable attention. Among the significant open challenges in this area of research is the question of scalability.
Cornerstone techniques such as Graph Convolutional Networks (GCNs)~\cite{Kipf2017} make the training dependent on the neighborhood of any given node. Since in many real-world graphs the number of neighbors grows exponentially with the number of hops taken, the scalability of such methods is a significant challenge. In recent years, several techniques have been proposed to make GCNs more scalable, including layer-wise sampling~\cite{Hamilton2018} and subgraph sampling~\cite{Chiang2019} approaches (see~\cref{sec:related}).

We contribute to this line of work by proposing a decentralised divide-and-conquer approach to improve the scalability of network embedding techniques. Our ``{\ltg}'' approach proceeds by first dividing the network into overlapping subgraphs (or ``patches'') and training separate local node embeddings for each patch (local in the sense that each patch is embedded into its own local coordinate system). The resulting local patch node embeddings are then transformed into a global node embedding (i.e. all nodes embedded into a single global coordinate system) by estimating a rigid motion applied to each patch using the As-Synchronized-As-Possible (ASAP) algorithm~\cite{Cucuringu2012a,Cucuringu2012b}. A key distinguishing feature of this ``decentralised'' approach is that we can train the different patch embeddings separately, without the need to keep parameters synchronised. The benefit of {\ltg} is threefold: (1) it is highly parallelisable as each patch is trained independently; (2) it can be used in privacy-preserving applications and federated learning
setups, where frequent communication between devices is often a limiting factor~\cite{Kairouz2021},
or ``decentralized'' organizations, where one needs to simultaneously consider data sets from different departments; (3) it can reflect varying structure across a graph through asynchronous parameter learning. Another important advantage of our {\ltg} approach is that it can be directly applied to improve the scalability of a large variety of network embedding techniques~\cite{Goyal2018a}, 
unlike most of the existing approaches reviewed in~\cref{sec:related} which are restricted to GCNs.

\section{Related work}
\label{sec:related}

The key scalability problems for GCNs only concern deep architectures where we have $l$ nested GCN layers. In particular, a single-layer GCN is easy to train in a scalable manner using mini-batch stochastic gradient descent (SGD). For simplicity, assume that we have a fixed feature dimension $d$, i.e., $d_\text{in}=d_\text{out}=d$ for all layers. The original GCN paper~\cite{Kipf2017} uses full-batch gradient descent to train the model which entails the computation of the gradient for all nodes before updating the model parameters. This is efficient in terms of time complexity per epoch ($O(lmd + lnd^2)$) where $n$ is the number of nodes and $m$ is the number of edges. However, it requires storing all the intermediate embeddings and thus has memory complexity $O(lnd + ld^2)$. Further, as there is only a single parameter update per epoch, convergence tends to be slow.

The problem with applying vanilla mini-batch SGD (where we only compute the gradient for a sample of nodes, i.e., the batch) to a deep GCN model is that the embedding of the nodes in the final layer depends on the embedding of all the neighbours of the nodes in the previous layer and so on iteratively. Therefore the time complexity for a single mini-batch update approaches that for a full-batch update as the number of layers increases, unless the network has disconnected components. There are mainly three families of methods~\cite{Chen2020,Chiang2019} that have been proposed to make mini-batch SGD training more efficient for GCNs.

\begin{description}[leftmargin=0pt, itemsep=\topsep]
\item[Layer-wise sampling.]
The idea behind layer-wise sampling is to sample a set of nodes for each layer of the nested GCN model and compute the embedding for sampled nodes in a given layer only based on embeddings of sampled nodes in the previous layer rather than considering all the neighbours as would be the case for vanilla SGD. This seems to have first been used by GraphSAGE~\cite{Hamilton2018}, where a fixed number of neighbours is sampled for each node at each layer. However, this results in a computational complexity that is exponential in the number of layers and also redundant computations as the same intermediate nodes may be sampled starting from different nodes in the batch. Later methods avoid the exponential complexity by first sampling a fixed number of nodes for each layer either independently (FastGCN~\cite{Chen2018a}) or conditional on being connected to sampled nodes in the previous layer (LADIES~\cite{Zou2019}) and reusing embeddings. Both methods use importance sampling to correct for bias introduced by non-uniform node-sampling distributions. Also notable is \cite{Chen2018b}, which uses variance reduction techniques to effectively train a GCN model using neighbourhood sampling as in GraphSAGE with only 2 neighbours per node. However, this is achieved by storing hidden embeddings for all nodes in all layers and thus has the same memory complexity as full-batch training.

\item[Linear model.]
Linear models remove the non-linearities between the different GCN layers which means that the model can be expressed as a single-layer GCN with a more complicated convolution operator and hence trained efficiently using mini-batch SGD. Common choices for the convolution operator are powers of the normalised adjacency matrix~\cite{Wu2019a} and variants of personalised Page-Rank (PPR) matrices~\cite{Busch2020,Chen2020,Bojchevski2020,Klicpera2019}. Another variant of this approach is \cite{frasca2020sign}, which proposes combining different convolution operators in a wide rather than deep architecture. There are different variants of the linear model architecture, depending on whether the non-linear feature transformation is applied before or after the propagation (see \cite{Busch2020} for a discussion), leading to predict-propagate and propagate-predict architectures respectively. The advantage of the propagate-predict architecture is that one can pre-compute the propagated node features (e.g., using an efficient push-based algorithm~\cite{Chen2020}) which can make training highly scalable. The disadvantage is that this will densify sparse features which can make training harder~\cite{Bojchevski2020}. However, the results from \cite{Busch2020} suggest that there is usually not much difference in prediction performance between these options (or the combined architecture where trainable transformations are applied before and after propagation).  

\item[Subgraph sampling.]
Subgraph sampling techniques~\cite{Zeng2019,Chiang2019,Zeng2020} construct batches by sampling an induced subgraph of the full graph. In particular, for subgraph sampling methods, the sampled nodes in each layer of the model in a batch are the same. In practice, subgraph sampling seems to outperform layer-wise sampling~\cite{Chen2020}. GraphSAINT~\cite{Zeng2020}, which uses a random-walk sampler with an importance sampling correction similar to \cite{Chen2018a,Zou2019}, seems to have the best performance so far. 
Our {\ltg} approach shares similarities with subgraph sampling, most notably ClusterGCN~\cite{Chiang2019}, which uses graph clustering techniques to sample the batches. The key distinguishing feature of our approach is that we train independent models for each patch whereas for ClusterGCN, model parameters have to be kept in sync for different batches, which hinders fully distributed training and its associated key benefits (see~\cref{introduction}).
\end{description}

\section{LOCAL2GLOBAL algorithm}\label{sec:l2g}

The key idea behind the {\ltg} approach to graph embedding is to embed different parts of a graph independently by splitting the graph into overlapping ``patches'' and then stitching the patch node embeddings together to obtain a global node embedding. 
The stitching of the patch node embeddings proceeds by estimating the rotations/reflections and translations for the embedding patches that best aligns them based on the overlapping nodes.

Consider a graph $G(V, E)$ with node set $V$ and edge set $E$. The input for the {\ltg} algorithm is a patch graph $G_p(\set{P}, E_p)$, where each node (i.e., a ``patch'') of the patch graph is a subset of $V$ and each patch $P_k \in \set{P}$ is associated with an embedding $\mat{X}^{(k)} \in \R^{|P_k|\times d}$. 
We require that the set of patches $\set{P}=\{P_k\}_{k=1}^p$ is a cover of the node set $V$ (i.e., $\bigcup_{k=1}^p P_k = V$), and that the patch embeddings all have the same dimension $d$. We further assume that the patch graph is connected and that
the patch edges satisfy the minimum overlap condition $\{P_i, P_j\} \in E_p \implies |P_i \cap P_j| \geq d+1$. 
Note that a pair of patches that satisfies the minimum overlap condition is not necessarily connected in the patch graph.

The {\ltg} algorithm for aligning the patch embeddings proceeds in two stages and is an evolution of the approach in \cite{Cucuringu2012a,Cucuringu2012b}. We assume that each patch embedding $\mat{X}^{(k)}$ is a perturbed part of an underlying global node embedding $\mat{X}$, where the perturbation is composed of reflection ($Z_2$), rotation (SO($d$)), translation ($\mathbb{R}^d$), and noise. The goal is to estimate the transformation applied to each patch using only pairwise noisy measurements of the relative transformation for pairs of connected patches. 
In the first stage, we estimate the orthogonal transformation to apply to each patch embedding, using a variant of the eigenvector synchronisation method \cite{Singer2011,Cucuringu2012a,Cucuringu2012b}. 
In the second stage, we estimate the patch translations by solving a least-squares problem. 
Note that unlike \cite{Cucuringu2012a,Cucuringu2012b}, we solve for translations at the patch level rather than solving a least squares problem for the node coordinates. 
This means that the computational cost for computing the patch alignment is independent of the size of the original network and depends only on the amount of patch overlap, the number of patches and the embedding dimension.

\subsection{Eigenvector synchronisation over orthogonal transformations} 

We assume that to each patch $P_i$, there corresponds an unknown group element $S_i \in O(d) \simeq Z_2 \times SO(d)$ (represented by a $d \times d$ orthogonal matrix), and for each pair of connected patches $(P_i, P_j) \in E_p$ we have a noisy proxy for $S_i S_j^{-1}$, which is precisely the setup of the \textit{group synchronization} problem.

For a pair of connected patches $P_i, P_j \in \set{P}$ such that $\{P_i, P_j\} \in E_p$ we can estimate the relative  rotation/reflection by applying the method from \cite{Horn1988}\footnote{Note that the roation/reflection can be estimated without knowing the relative translation.} to their overlap as $|P_i \cap P_j| \geq d+1$.
Thus, we can construct a block matrix $\mat{R}$ where $\mat{R}_{ij}$ is the $d \times d$ orthogonal matrix representing the estimated relative transformation from patch $P_j$ to patch $P_i$ if $\{P_i, P_j\} \in E_p$ and $\mat{R}_{ij}=\mat{0}$ otherwise, such that $\mat{R}_{ij} \approx \mat{S}_i \mat{S}_j^{T}$ for connected patches.

In the noise-free case, we have the consistency equations 
$\mat{S}_i = \mat{R}_{ij} \mat{S}_j$
for all $i, j$ such that $\{P_i, P_j\} \in E_p$. 
We can combine the consistency equations for all neighbours of a patch to get
\begin{equation}
\mat{S}_i = \mat{M}_{ij}\mat{S}_j, \qquad \mat{M}_{ij} = \frac{\sum_{j} w_{ij} \mat{R}_{ij}}{\sum_{j} w_{ij}},
\label{eq:syn_sum}
\end{equation}
where we use $w_{ij} = |P_i \cap P_j|$ to weight the contributions as we expect a larger overlap to give a more robust estimate of the relative transformation. 
We can write \cref{eq:syn_sum} as 
$\mat{S} = \mat{M} \mat{S}$, where $\mat{S} = (\mat{S}_1, \ldots, \mat{S}_p)^T$ is a $pd \times d$ block-matrix and $\mat{M}$ is a $pd \times pd$ block-matrix. 
Thus, in the noise-free case, the columns of $\mat{S}$ are eigenvectors of $\mat{M}$ with eigenvalue 1. Thus, following \cite{Cucuringu2012a,Cucuringu2012b}, we can use the $d$ leading eigenvectors\footnote{While $\mat{M}$ is not symmetric, it is similar to a symmetric matrix and thus admits a basis  of real, orthogonal eigenvectors.} of $\mat{M}$ as the basis for estimating the transformations. 
Let $\mat{U}= (\mat{U}_1, \ldots, \mat{U}_p)^T$ be the $pd \times d$ matrix whose columns are the $d$ leading eigenvectors of $\mat{M}$, where $\mat{U}_i$ is the $d \times d$ block of $\mat{U}$ corresponding to patch $P_i$. We obtain the estimate $\hat{\mat{S}}_i$ of $\mat{S}_i$ by finding the nearest orthogonal transformation to $\mat{U}_i$ using an SVD \cite{Horn1988}, and hence the estimated rotation-synchronised embedding of patch $P_i$ is $\hat{\mat{X}}^{(i)} = \mat{X}^{(i)} \hat{\mat{S}}_i^T$.

\subsection{Synchronisation over translations}

After synchronising the rotation of the patches, we can estimate the translations by solving a least squares problem. 
Let $\hat{\mat{X}}_i^{(k)} \in \R^d$ be the (rotation-synchronised) embedding of node $i$ in patch $P_k$ ($\hat{\mat{X}}_i^{(k)}$ is only defined if $i\in P_k$). 
Let $\mat{T}_k\in \R^d$ be the translation of patch $k$, then in the noise-free case we have the consistency equations
\begin{equation}
\hat{\mat{X}}_i^{(k)} + \mat{T}_k = \hat{\mat{X}}_i^{(l)} + \mat{T}_l,\qquad i \in P_k \cap P_l .\label{eq:consistency_single}
\end{equation}
We can combine the conditions in \cref{eq:consistency_single} for each edge in the patch graph to obtain
\begin{equation}
\mat{B} \mat{T} = \mat{C}, \qquad \mat{C}_{(P_k, P_l)} = \frac{\sum_{i \in P_k \cap P_l} \hat{\mat{X}}_i^{(k)} - \hat{\mat{X}}_i^{(l)}}{|P_k \cap P_l|},  \label{eq:consistency_combined}
\end{equation}
where $\mat{T} \in \R^{|\set{P}| \times d}$ is the matrix such that the $k$th row of $\mat{T}$ is the translation $\mat{T}_k$ of patch $P_k$ and $\mat{B}\in \{-1,1\}^{|E_p|\times |\set{P}|}$ is the incidence matrix of the patch graph with entries $\mat{B}_{(P_k,P_l),j} = \delta_{lj}-\delta_{kj}$, where $\delta_{ij}$ denotes the Kronecker delta. 
\Cref{eq:consistency_combined} defines an overdetermined linear system that has the true patch translations as a solution in the noise-free case. 
In the practical case of noisy patch embeddings, we can instead solve \cref{eq:consistency_combined} in the least-squares sense
\begin{equation}
\hat{\mat{T}} = \argmin_{\mat{T} \in \R^{p\times d}} \norm{\mat{B}\mat{T} - \mat{C}}_2^2.
\label{eq:translation}
\end{equation}
We estimate the aligned node embedding $\bar{\mat{X}}$ in a final step using the centroid of the aligned patch embeddings of a node, i.e.,
\[
\bar{\mat{X}}_i = \frac{\sum_{\{P_k \in \set{P} \given i \in P_{k}\}} \hat{\mat{X}}_i^{(k)}+\hat{\mat{T}}_k}{|\{P_k \in \set{P} \given i \in P_{k}\}|}.
\]

\subsection{Scalability of the {\ltg} algorithm}

The patch alignment step of {\ltg} is highly scalable and does not directly depend on the size of the input data. The cost for computing the matrix $M$ is $O(|E_p|od^2)$ where $o$ is the average overlap between connected patches (typically $o \sim d$) and the cost for computing the vector $b$ is $O(|E_p|od)$. Both operations are trivially parallelisable over patch edges. The translation problem can be solved with an iterative least-squares solver with a per-iteration complexity of $O(|E_p|d)$.
The limiting step for {\ltg} is usually the synchronisation over orthogonal transformations which requires finding $d$ eigenvectors of a $d|\set{P}| \times d|\set{P}|$ sparse matrix with $|E_p|d^2$ non-zero entries for a per-iteration complexity of $O(|E_p|d^3)$. This means that in the typical scenario where we want to keep the patch size constant, the patch alignment scales almost linearly with the number of nodes in the dataset, as we can ensure that the patch graph remains sparse, such that $|E_p|$ scales almost linearly with the number of patches. The $O(|E_p|d^3)$ scaling puts some limitations on the embedding dimension attainable with the {\ltg} approach, though, as we can see from the experiments  in \cref{sec:embeddings}, it remains feasible for reasonably high embedding dimension.

The preprocessing to divide the network into patches scales as $O(m)$. The speed-up attainable due to training patches in parallel depends on the oversampling ratio (i.e., the total number of edges in all patches divided by the number of edges in the original graph). As seen in \cref{sec:embeddings}, we achieve good results with moderate oversampling ratios. 

\section{Experiments}

\subsection{Data sets}

We consider two data sets to test the viability of the {\ltg} approach to graph embeddings, the Cora citation data set from \cite{Yang2016} and the Amazon photo data set from \cite{Shchur2019}. We consider only nodes and edges in the largest connected component (LCC). We show some statistics of the data sets in \cref{tab:data}.
\vspace{-1mm}
\begin{table}[H]
\vspace{-1mm} 
\begin{tabular}{r|S[table-number-alignment = right]S[table-number-alignment = right]S[table-number-alignment = right]}
& {nodes in LCC} & {edges in LCC} & {features}\\
\hline
Cora & 2485 & 10138 & 1433\\
Amazon photo & 7487 & 238086 & 745
\end{tabular}
\caption{Data sets\label{tab:data}}
\vspace{-\textfloatsep} % remove excessive space after table
\end{table}
\vspace{-1mm}

\begin{algorithm}[tb]
\caption{Sparsify patch graph\label{alg:sparsify}}
\KwIn{$G_p(\set{P}, E_p)$, $G(V, E)$, target patch degree $k$}
\KwResult{sparsified patch graph $G_p(\set{P}, \tilde{E}_p)$}
\ForEach{$\{P_i, P_j\} \in E_p$}{
Compute conductance weight
\begin{algomathdisplay}
c_{ij} = \tfrac{|\{(u, v) \in E \given u \in P_i, v \in P_j\}|}{\min(|\{(u, v) \in E \given u \in P_i\}|, |\{(u, v) \in E \given u \in P_j\}|)}
\end{algomathdisplay}
}
\ForEach{$\{P_i, P_j\} \in E_p$}{
Compute effective resistance $r_{ij}$ between $P_i$ and $P_j$ in $G_p(\set{P}, E_p, c)$ using the algorithm of \cite{Spielman2011a}\;
Let $w_{ij} = r_{ij} c_{ij}$\;
}
Initialize $\tilde{E}_p$ with a maximum spanning tree of $G_p(\set{P}, E_p, w)$\;
Sample the remaining $(k-1)p + 1$ edges from $E_p \setminus \tilde{E}_p$ without replacement and add them to $\tilde{E}_p$, where edge $\{P_i, P_j\}$ is sampled with probability $w_{ij}$\;
\Return $G_p(\set{P}, \tilde{E}_p)$
\end{algorithm}

\begin{algorithm}[tb]
\caption{Create overlapping patches\label{alg:overlap}}
\KwIn{$\set{C}$, $E_p$, $G(V, E)$, min overlap $l$, max overlap $u$}
\KwResult{Overlapping patches $\set{P}$}
Initialise $\set{P} = \set{C}$\;
Define the neighbourhood of a set of nodes $U$ as
\begin{algomathdisplay}
N(U) = \{j \given i \in U, (i, j) \in E, j \notin U\}
\end{algomathdisplay}
\ForEach{$P_i \in \set{P}$}{
\ForEach{$P_j$ s.t. $\{P_i, P_j\} \in E_p$}{
	Let $F = N(C_i) \cap C_j$\;
	\While{$|P_i \cap C_j| < l/2$}{
		\If{$|F| + |P_i \cap C_j| > u/2$}{
			reduce $F$ by sampling uniformly at random such that $|F| = u/2 - |P_i \cap C_j|$\;}
		Let $P_i = P_i \cup F$\;
		Let $F = (N(F) \cap C_j) \setminus P_i$\;
		}
	}
}
\Return{$\set{P}$}
\end{algorithm}

\vspace{-2mm} 
\subsection{Patch graph construction}
The first step in the {\ltg} embedding pipeline is to divide the network $G(V, E)$ into overlapping patches. 
In some federated-learning applications, the network may already be partitioned and some or all of the following steps may be skipped provided the resulting patch graph is connected and satisfies the minimum overlap condition for the desired embedding dimension. 
Otherwise, we proceed by first partitioning the network into non-overlapping clusters and then enlarging clusters to create overlapping patches. 
This two-step process makes it easier to ensure that patch overlaps satisfy the conditions for the {\ltg} algorithm without introducing excessive overlaps than if we were to use a clustering algorithm that produces overlapping clusters directly. We use the following pipeline to create the patches:
\begin{itemize}
\item \textbf{Partition the network} into $p$ non-overlapping clusters $\set{C}=\{C_k\}_{k=1}^p$ such that $|C_k| \geq \frac{d+1}{2}$ for all $k$. We use METIS~\cite{Karypis1998} to cluster the networks for the experiments in \cref{sec:embeddings}. However, for very large networks, more scalable clustering algorithms such as FENNEL~\cite{Tsourakakis2014} could be used.
\item \textbf{Initialize the patches} to $\set{P} = \set{C}$ and define the patch graph $G_p(\set{P}, E_p)$, where $\{P_i, P_j\} \in E_p$ iff there exist nodes $i \in P_i$ and $j \in P_j$ such that $\{i, j\} \in E$. (Note that if $G$ is connected, $G_p$ is also connected.)
\item \textbf{Sparsify the patch graph} $G_p$ to have mean degree $k$ using \cref{alg:sparsify} adapted from the effective-resistance sampling algorithm of \cite{Spielman2011a}. 
\item \textbf{Expand the patches} to create the desired patch overlaps. We define a lower bound $l \geq d+1$ and upper bound $u$ for the desired patch overlaps and use \cref{alg:overlap} to expand the patches such that $|P_i \cap P_j| \geq l$ for all $\{P_i, P_j\} \in E_p$.
\end{itemize}

For Cora, we split the network into 10 patches and sparsify the patch graph to a target mean degree $k=4$. We set the lower bound for the overlap to $l=129$ and upper bound to $u=256$. For Amazon photo, we split the network into 20 patches and sparsify the patch graph to a target mean degree of $k=5$. We set the lower bound for the overlap to $l=256$ and the upper bound to $u=512$.

\subsection{Embedding model}
As embedding method we consider the variational graph auto-encoder (VGAE) architecture of~\cite{Kipf2016}. We use the Adam optimizer~\cite{Kingma2015} for training with learning rate set to 0.01 for Cora and 0.001 for Amazon photo and train all models for 200 epochs. We set the hidden dimension of the models to $2\times d$ for Cora and to $4\times d$ for Amazon photo where $d$ is the embedding dimension.

\vspace{-1mm} 
\subsection{Results}
\label{sec:embeddings}

\begin{figure}
\begin{minipage}{\linewidth}
\includegraphics[width=\linewidth, clip, trim={0 0.2cm 0 0.8cm}]{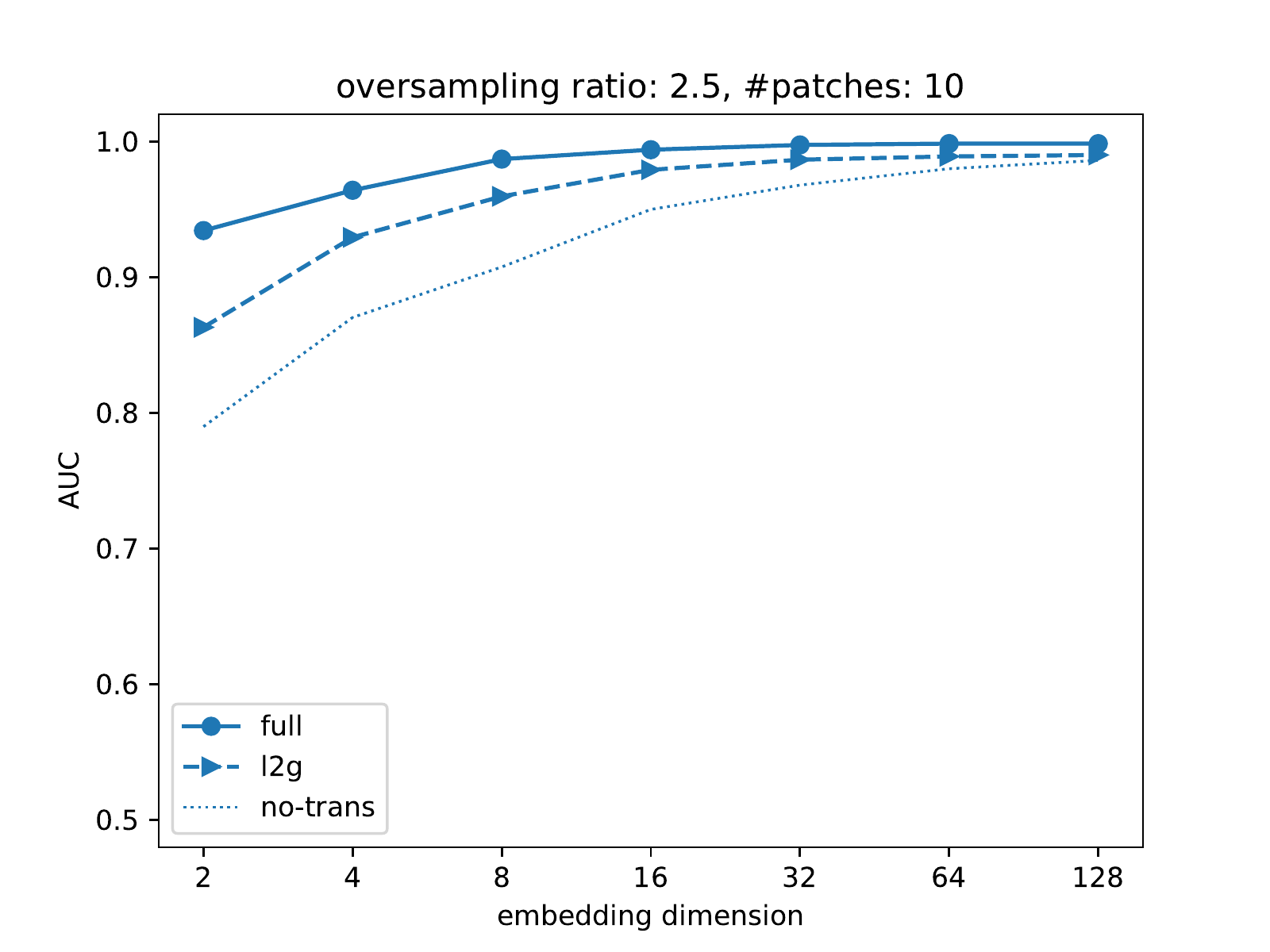}
\subcaption{Cora\label{fig:auc_scores:cora}}
\end{minipage}\par\medskip

\begin{minipage}{\linewidth}
\includegraphics[width=\linewidth, clip, trim={0 0.2cm 0 0.8cm}]{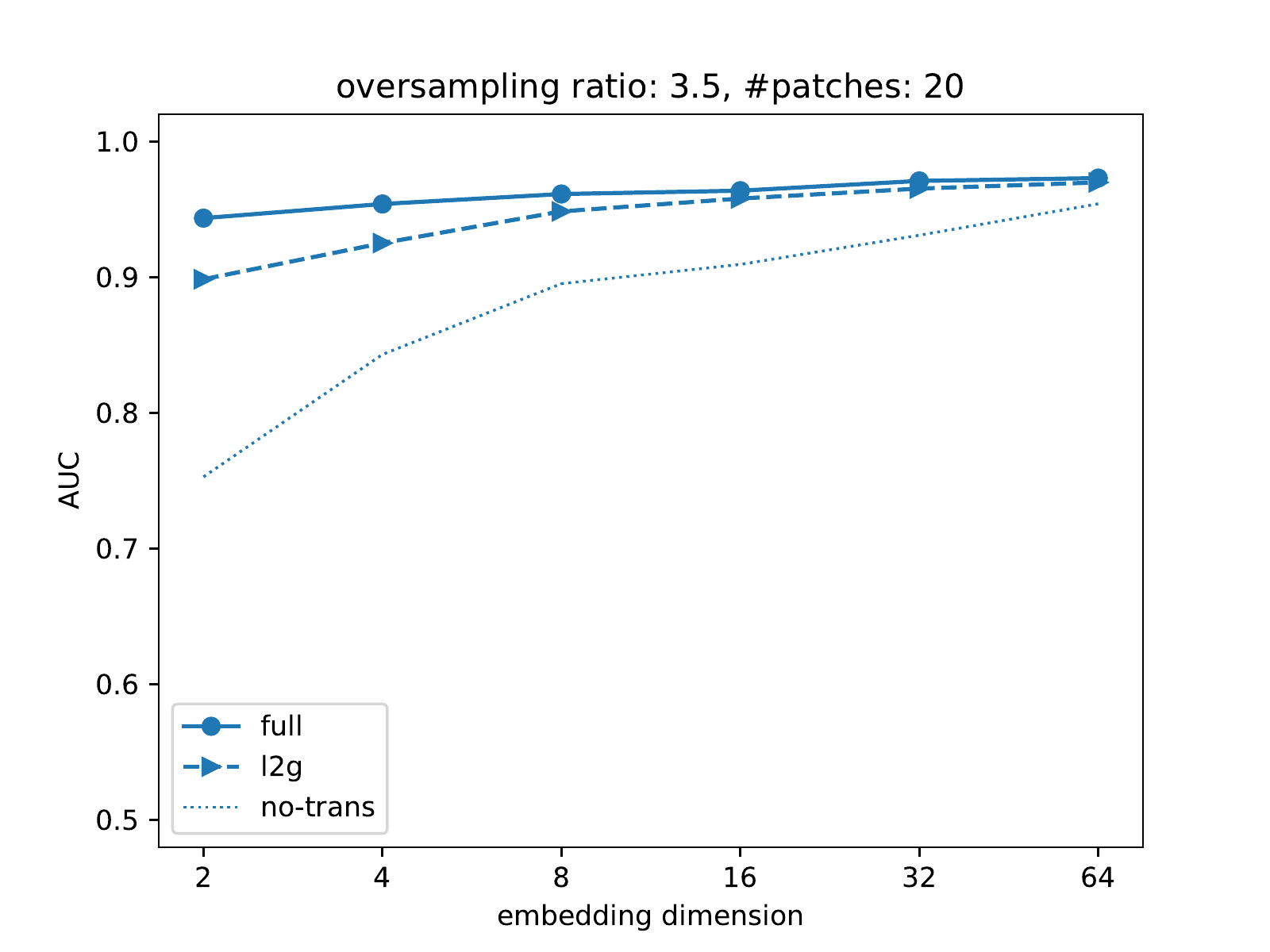}%
\subcaption{Amazon photo\label{fig:auc_scores:amz_photo}}
\end{minipage}
\caption{AUC network reconstruction score as function of embedding dimension using full data or stitched patch embeddings for \subref{fig:auc_scores:cora} Cora and \subref{fig:auc_scores:amz_photo} Amazon photo. \label{fig:auc_scores}}
\end{figure}

As a first test case for the viability of the {\ltg} approach, we consider a network reconstruction task. 
We train the models using all edges in the largest connected component and compare three training scenarios
\begin{description}[leftmargin=\parindent]
\item[full:] Model trained on the full data.
\item[l2g:] Separate models trained on the subgraph induced by each patch and stitched using the {\ltg} algorithm.
\item[no-trans:] Same training as l2g but node embeddings are obtained by taking the centroid over patch embeddings that contain the node without applying the alignment transformations. 
\end{description}
We evaluate the network reconstruction error using the AUC scores based on all edges in the largest connected component as positive examples and the same number of randomly sampled non-edges as negative examples. We train the models for 200 epochs using full-batch gradient descent. We show the results in \cref{fig:auc_scores}. For `full', we report the best result out of 10 training runs. For `l2g' and `no-trans', we first identify the best model out of 10 training runs on each patch and report the results for stitching the best models. 

Overall, the gap between the results for `l2g` and `full` is small and essentially vanishes for higher embedding dimensions. The aligned `l2g` embeddings consistently outperform the unaligned `no-trans' baseline.

\vspace{-1mm} 
\section{Conclusion}
In this work, we introduced a framework that can significantly improve the computational scalability of generic graph embedding methods, rendering them scalable to real-world applications that involve massive graphs, potentially with millions or even billions of nodes.  At the heart of our pipeline is the {\ltg} algorithm, a divide-and-conquer approach that first decomposes the input graph into overlapping clusters (using one's method of choice), computes entirely local embeddings via the  preferred embedding method, for each resulting cluster (exclusively using information available at the nodes within the cluster), and finally stitches the resulting local embeddings into a globally consistent embedding, using established  machinery from the group synchronization literature. 

Our preliminary results on medium-scale data sets are promising and achieve comparable accuracy on graph reconstruction as globally trained VGAE embeddings. Our ongoing work consists of two keys steps. A first is to further demonstrate the scalability benefits of {\ltg} on large-scale data sets using a variety of embedding techniques and downstream tasks by comparing with state-of-the-art synchronised subgraph sampling methods, as well as exploring the trade-off between parallelisability and embedding quality as a function of patch size and overlap.
A second is to demonstrate particular benefits of locality and asynchronous parameter learning. These have clear advantages for privacy preserving and federated learning setups. It would also be particularly interesting to assess the extent to which this 
{\ltg}  approach can outperform global methods. The intuition and hope in this direction stems from the fact that asynchronous locality can be construed as a regularizer (much like sub-sampling, and similar to dropout) and could potentially lead to better generalization and alleviate the oversmoothing issues of deep GCNs, as observed in \cite{Chiang2019}.

\bibliographystyle{ACM-Reference-Format}
\bibliography{bibliography.bib}

\end{document}